\documentclass[10pt,twocolumn,letterpaper]{article}

\usepackage[pagenumbers]{cvpr} % To force page numbers, e.g. for an arXiv version

\usepackage[table,dvipsnames]{xcolor}

\usepackage{graphicx}
\usepackage{multirow}
\usepackage{mypythonhighlight}
\usepackage{algorithm}
\usepackage{algorithmic}
\usepackage[accsupp]{axessibility}

\usepackage{tcolorbox}
\tcbset{colback=white}
\colorlet{titleblue}{blue!80!black}
\colorlet{titlered}{red!80!black}
\colorlet{titlegreen}{green!80!black}
\colorlet{darkgreen}{green!50!black}
\newcommand{\inc}[1]{{\color{darkgreen}#1}}
\newcommand{\dec}[1]{{\color{titlered}#1}}
\newcommand{\TBlack}[1]{\textnormal{\ttfamily\color{black}#1}\unskip}
\newcommand{\TBlue}[1]{\textnormal{\ttfamily\color{titleblue}#1}\unskip}

\newcommand{\pilot}{LaMPilot\xspace}
\newcommand{\pilotb}{LaMPilot-Bench\xspace}
\newcommand{\Imbf}{\mathbf{I}}
\newcommand{\Ambf}{\mathbf{A}}
\newcommand{\Cmbf}{\mathbf{C}}
\newcommand{\Pmbf}{\mathbf{P}}
\newcommand{\Rmbf}{\mathbf{R}}
\newcommand{\Fmbf}{\mathbf{F}}

\definecolor{cvprblue}{rgb}{0.21,0.49,0.74}
\usepackage[pagebackref,breaklinks,colorlinks,citecolor=cvprblue]{hyperref}

\def\paperID{7857} % *** Enter the Paper ID here
\def\confName{CVPR}
\def\confYear{2024}

\title{LaMPilot: An Open Benchmark Dataset for Autonomous Driving with Language Model Programs}

\author{%
  Yunsheng Ma$^1$\thanks{Equal Contribution} , Can Cui$^1$\footnotemark[1] , Xu Cao$^2$\footnotemark[1] , Wenqian Ye$^3$, Peiran Liu$^{1}$, Juanwu Lu$^{1}$, \\ Amr Abdelraouf$^4$, Rohit Gupta$^4$, Kyungtae Han$^4$, Aniket Bera$^1$, James M. Rehg$^{2}$, Ziran Wang$^{1}$\\
  $^{1}$Purdue University\quad
  $^{2}$University of Illinois Urbana-Champaign\\
  $^{3}$University of Virginia\quad 
  $^{4}$InfoTech Labs, Toyota Motor North America \\
  \texttt{\small \{yunsheng, ziran\}@purdue.edu}
}

\begin{document}
\maketitle
Autonomous driving (AD) has made significant strides in recent years. However, existing frameworks struggle to interpret and execute spontaneous user instructions, such as ``overtake the car ahead." Large Language Models (LLMs) have demonstrated impressive reasoning capabilities showing potential to bridge this gap. In this paper, we present \pilot, a novel framework that integrates LLMs into AD systems, enabling them to follow user instructions by generating code that leverages established functional primitives. We also introduce \pilotb, the first benchmark dataset specifically designed to quantitatively evaluate the efficacy of language model programs in AD. Adopting the \pilot framework, we conduct extensive experiments to assess the performance of off-the-shelf LLMs on \pilotb. Our results demonstrate the potential of LLMs in handling diverse driving scenarios and following user instructions in driving. To facilitate further research in this area, we release our code and data at \href{https://github.com/PurdueDigitalTwin/LaMPilot}{GitHub.com/PurdueDigitalTwin/LaMPilot}.
\section{Introduction}
\label{sec:intro}
Autonomous driving (AD) has witnessed remarkable progress in recent years, with an increasing number of commercial autonomous vehicles (AVs) being deployed on public roads~\cite{mckinsey_center_for_future_mobility_autonomous_2023}. State-of-the-art AD systems can be broadly classified into two categories: 1) a modular approach, where standalone models are developed for perception, prediction, and planning independently~\cite{grigorescu_survey_2020}, and 2) an end-to-end approach that directly maps sensor data to control signals via a single neural network~\cite{jiang_vad_2023,hu_planning-oriented_2023,shao_reasonnet_2023}. Despite significant breakthroughs, both approaches struggle to handle arbitrary user commands effectively, such as ``overtake the car in front of me."

Large Language Models (LLMs) have demonstrated impressive capabilities in language comprehension and reasoning~\cite{sun_survey_2024,yao_react_2023}, showing potential to improve the safety, explainability, and user-friendliness of AVs. Utilizing LLMs to solve AV-related tasks is gaining momentum~\cite{cui_survey_2024,zhou_vision_2023,yang_llm4drive_2023,li_towards_2023,gao_survey_2024}. However, the integration of LLMs into existing AD frameworks presents several challenges. Firstly, there is a lack of well-established paradigms for incorporating LLMs into the decision-making process of AVs. Secondly, there is a shortage of benchmarks designed to evaluate and compare the performance of LLM-based agents in the context of driving.

To address these challenges, we propose a novel framework called \pilot. Inspired by Code as Policy~\cite{liang_code_2023}, which utilizes code LLMs to write robot policy code, \pilot employs Language Model Programs (LMPs) as the action space instead of low-level vehicle control signals. We equip LLMs with APIs that cover various functional primitives, enabling them to connect natural language instructions to executable driving plans through code generation.

We also introduce \pilotb, the first benchmark for evaluating LMPs in AD. The primary objective for agents operating within \pilotb is to accomplish assigned tasks safely and efficiently. \pilotb incorporates an interactive simulator and evaluator, featuring programmatic scoring mechanisms to assess policy performance. The main contributions of this work are summarized as follows:

\begin{itemize}
\item We propose \pilot, a novel framework that integrates LLMs into autonomous driving systems, enhancing their ability to interpret and follow user commands.
\item We introduce \pilotb, the first benchmark dataset designed to evaluate the performance of LLM-powered agents in autonomous driving. Each scenario in \pilotb consists of a task described in natural language, along with a simulated environment for comprehensive evaluation.
\item Adopting the \pilot framework, we conduct extensive experiments to assess the performance of off-the-shelf LLMs on \pilotb.  Our results demonstrate the great potential of LLMs in handling diverse driving scenarios and following user instructions.
\end{itemize}

\begin{figure*}[!t]
    \centering
    \includegraphics[width=0.98\linewidth]{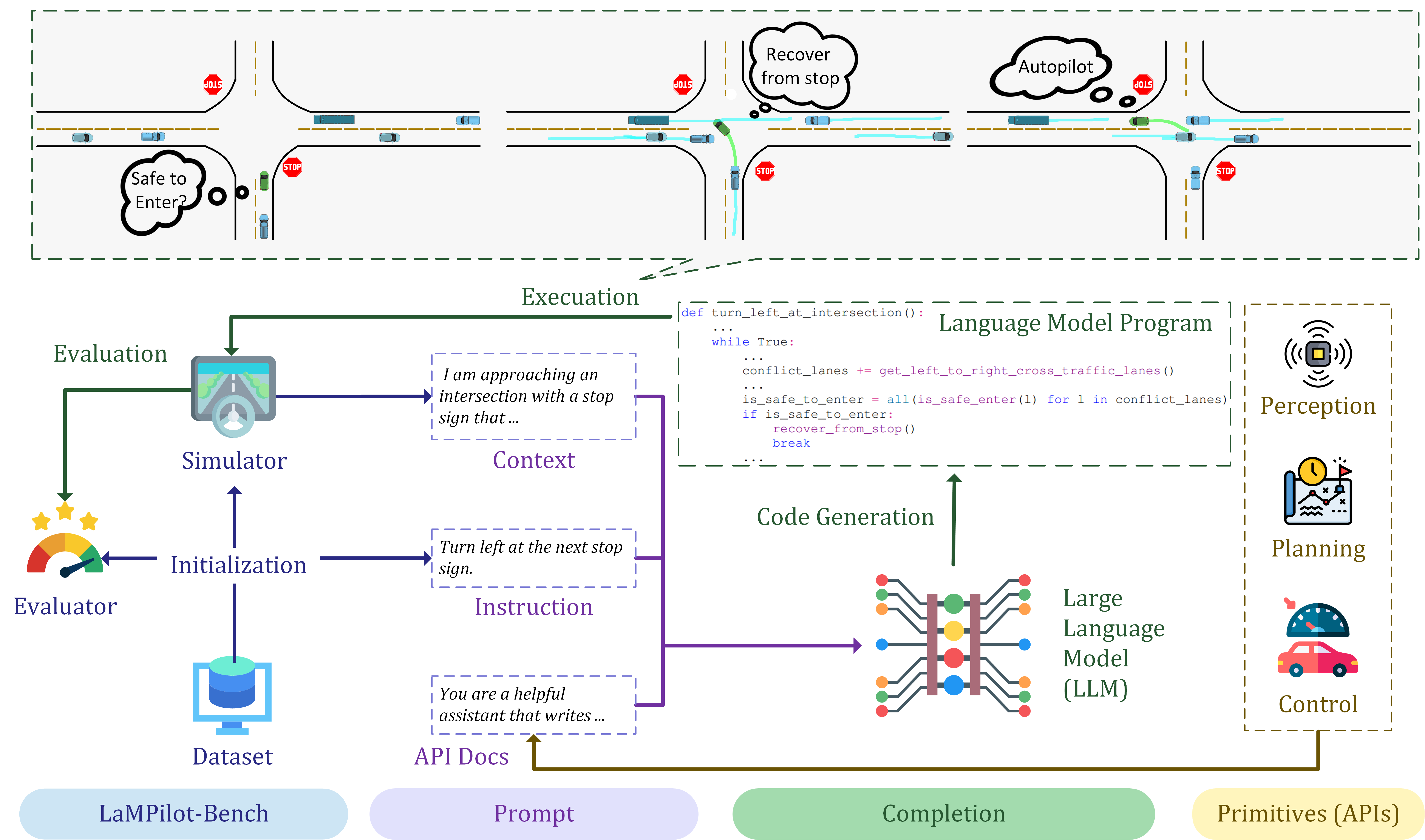}
    \caption{An overview of the \pilot framework. The Large Language Model (LLM) receives a prompt containing human instructions, driving context, and API documentation. It then writes language model programs that serve as driving policies. These policies are executed in the simulator to complete the specified driving task and are subsequently evaluated by the evaluator to assess the effectiveness of the generated policy code.}
    \vspace{-2mm}
    \label{fig:framework}
\end{figure*}

%routes_testing_03_02_23_44_23 0155
\begin{table*}[t!]\centering
\begin{minipage}{1.0\linewidth}\vspace{0mm}    
\centering
\begin{tcolorbox} 
\centering
\footnotesize
\begin{tabular}{p{0.97\columnwidth} c}
\TBlack{ {\bf API Docs ($\mathbf{A}$)} } &\\
You are a helpful assistant that writes python code to complete any autonomous driving task specified by me. \\
Here are the APIs you can use: ... \textbf{[omitted]} \\
% I will give you \\
% Command: ... \\
% Context: ... \\
% You should then respond to me with \\
% Plan: How to complete the task step by step. \\
% Code: \\
%     1) Write a function taking NO argument. \\
%     2) Only use the APIs that I provided. \\
% ... [omitted]
\TBlack{ {\bf Instruction ($\mathbf{I}$)} } &\\
Go around the car in front of you using the right lane.\\
\TBlack{ {\bf Driving Context ($\mathbf{C}$)} } &\\ 
My current speed is 31.1 m/s. I am driving on a highway with 2 lanes in my direction, and I am in the 2nd lane from the right. There is a car in front of me in my lane, at a distance of 44.9 m, with a speed of 31.1 m/s.

\hrulefill & \\

\TBlue{{\bf Chain-of-Thought Reasoning ($\mathbf{R}$)}} &\\
1) Check if there is a right lane available.
2) Check if there is a vehicle in front of the ego vehicle in the current lane within 100 meters.
3) If so, check if it is safe to enter the right lane.
4) If it is safe to enter the right lane, set the target lane of the ego vehicle to the right lane.
5) Monitor the distance to the target vehicle.
6) If the target vehicle is still in front of the ego vehicle, set the target speed of the ego vehicle to 1.5 times the speed of the target vehicle to overtake.
7) Repeat steps 5-6 until the target vehicle is no longer in front of the ego vehicle. \\
\TBlue{ {\bf Code as Policies ($\mathbf{P}$)} } &\\
\vspace{-2mm}
\begin{python}
def overtake_using_right_lane():
    ego_vehicle = get_ego_vehicle()
    current_lane = get_lane_of(ego_vehicle)
    right_lane = get_right_lane(ego_vehicle)
    target_vehicle = detect_front_vehicle_in(current_lane)
    if right_lane is None:
        say("There is no right lane to change into.")
        return
    if target_vehicle is None:
        say("There is no vehicle in front of me.")
        return
    # Check if it is safe to enter the right lane
    while True:
        if is_safe_enter(right_lane):
            set_target_lane(right_lane)
            break
        yield autopilot()
    # Monitor the distance to the target vehicle
    while True:
        distance_to_target = get_distance_between_vehicles(ego_vehicle, target_vehicle)
        if distance_to_target < 0:  # The target vehicle is still in front of the ego vehicle
            set_target_speed(get_speed_of(target_vehicle) * 1.5)
            yield autopilot()
        else:
            break
\end{python}
\end{tabular}
\vspace{-3mm}
\end{tcolorbox}
\caption{This example shows the chain-of-thought reasoning and code generation process of \pilot. Contents with \TBlue{blue} titles are generated by LLMs.}
\label{tab:full_example}
\end{minipage}
\vspace{-5mm}
\end{table*}
\section{Related Works}
\subsection{Language-Guided Driving}
Recent studies highlight the potential role of language in enhancing self-driving technology. Companies like Wayve are integrating natural language to improve the learning and explainability of their driving models. For example, their LINGO-1 system combines vision, language, and action modalities~\cite{wayve_technologies_ltd_lingo-1_2023,marcu_lingoqa_2023}.

Researchers have also been incorporating LLMs into AD systems for various purposes, such as generalization~\cite{jin_surrealdriver_2023,wen_dilu_2024,wang_drivemlm_2023,tian_drivevlm_2024,chen_driving_2024,mao_gpt-driver_2023,mao_language_2023,li_driving_2024,pan_vlp_2024,fu_drive_2024,wen_road_2023,cui_receive_2024,fu_limsim_2024,zhou_embodied_2024}, interpretability~\cite{xu_drivegpt4_2023,sha_languagempc_2023,yuan_rag-driver_2024,nie_reason2drive_2023,ding_hilm-d_2023}, and human-vehicle interaction~\cite{cui_large_2023,shao_lmdrive_2024,cui_drive_2024,wang_drivemlm_2023,sima_drivelm_2023,yang_human-centric_2024}. DiLu~\cite{wen_dilu_2024}, for instance, instills knowledge-driven capability into AD systems by considering how humans drive. It introduces a new framework that includes an interactive environment, a driver agent, and a memory component. LMDrive~\cite{shao_lmdrive_2024} proposes an end-to-end, language-based AD framework that processes camera-LiDAR sensor data, comprehends natural language driving instructions, and directly generates vehicle control signals.

However, most existing works either provide high-level goals, let LLMs generate the ego vehicle's future trajectories, or involve detailed vehicle control signals like throttle and steering angles. In contrast, our work uniquely uses LLMs to generate program code as driving policies, which are then executed with classical planners or controllers. Our approach takes advantage of the strengths of both LLMs and traditional AD components.

\subsection{Large Language Models for Code Generation}
The emergence of powerful LLMs has opened up new possibilities for leveraging programming languages to perform various tasks across multiple domains~\cite{yang_if_2024}. Code generation~\cite{zan_large_2023} is one of the most prominent applications, where LLMs are used to generate code snippets based on natural language descriptions. LLMs have also been employed for information extraction tasks using code as an intermediate representation~\cite{wang_code4struct_2023,wang_mustie_2023}. The robotics domain has also benefited from the integration of LLMs and code. Code as Policy~\cite{liang_code_2023}, ProgPrompt\cite{singh_progprompt_2023}, and VoxPoser~\cite{huang_voxposer_2023} demonstrate how LLMs can generate code to control robots and manipulate objects in 3D environments. In the vision domain, works such as ViperGPT~\cite{suris_vipergpt_2023} showcase the potential of using LLMs to generate code for visual processing and reasoning. Moreover, code provides a foundation for logical planning~\cite{wang_leti_2023,lu_chameleon_2023}, which can be leveraged by LLMs to tackle complex, long-horizon tasks. The combination of LLMs and code has revealed a new paradigm with broad applicability across numerous engineering and scientific domains.

\section{\pilot Methodology}
This section elaborates on the framework of \pilot, which integrates the world knowledge and reasoning capabilities of LLMs into an AD system to enable user instruction following. \pilot uses LLMs to generate LMPs that serve as the action space. LMPs have the inherent potential to symbolize actions that are both temporally generalizable and hierarchically structured. For instance, a complex maneuver like overtaking can be programmatically split into a series of actions, including lane changes, acceleration, and possibly another lane change, composed with conditional logic.

\pilot is designed as an add-on module for existing AD systems rather than as a standalone solution. It synergizes the emerging capabilities of LLMs with the proven efficiency of classical AD algorithms. Instead of directly issuing real-time, low-level control signals, the LLM generates code snippets at lower frequencies. These snippets, formulated as temporary policies, guide the strategic navigation of the ego vehicle based on user instructions. 

\subsection{Prompts}
As shown in \cref{fig:framework}, the \pilot framework takes inputs including API documentation ($\Ambf$), human instructions ($\Imbf$), and driving context ($\Cmbf$). It leverages Chain-of-Thought (CoT) reasoning~\cite{wei_chain--thought_2022} to write policy code ($\Pmbf$). This process can be formulated as:
\begin{equation}
    [\Rmbf, \Pmbf] = \ell([\Ambf, \Imbf, \Cmbf])
\end{equation}
where $\ell$ represents the LLM backbone.

An example to demonstrate the code generation process is provided in \cref{tab:full_example}. The instruction is a spontaneous command from the user. In addition, we also provide environment contexts and API documentation as part of the input prompt for the LLM.

The driving contexts include relevant information about the driving environment. We develop an interface that converts the numerical vector data in simulation into a narrative format using a structured language generator~\cite{chen_driving_2024}. This narrative provides a natural input interface for the LLM and does not require additional fine-tuning for interpretation.

\subsection{Functional Primitives}
\label{sec:api}
Policy code involves calls to functional primitives. They are implemented with classical planning and control algorithms. Due to the autoregressive nature of LLMs, generating long completions can introduce significant latency, making them less suitable for time-critical tasks like collision avoidance. The main idea is to utilize LLMs for strategic planning while intentionally avoiding their direct involvement in low-level control tasks. 

Inspired by the concept of Responsibility-Sensitive Safety~\cite{shalev-shwartz_formal_2017}, which ensures that the autonomous driving system will not issue a command that would lead to an accident, we include safety heuristics in our API implementations to minimize risks. Our API suite is categorized into four main types:
\begin{itemize}
    \item Ego APIs: They provide information about the status of the ego vehicle.
    \item Perception APIs: They include functions like object and lane detection, which can be called to acquire information about the surrounding environment. 
    \item Navigation APIs: They provide route planning functionalities given a target destination.
    \item Control APIs: They translate LLM-generated code into low-level control signals for the vehicle. 
\end{itemize}

In the event of an exception during policy execution, the ego vehicle automatically switches to a predefined \pyth{autopilot} mode to prevent undefined behaviors. To provide LLMs with essential information about the available APIs and guide their correct usage, we include the API Documentation ($\Ambf$) in the prompt. These documents include not only the input and output specifications but also provide descriptions of their usage and the underlying logic.

\subsection{Completions} 
The completion produced by the LLMs is anticipated to be valid functions, written with the provided APIs. These functions could range from straightforward, one-off functions to more complex generator functions. A simple example is demonstrated below, where the target speed of the ego vehicle is altered:

\begin{minipage}{\linewidth}
\begin{python}
def decrease_speed_by_5():
    set_target_speed(get_target_speed() - 5)
\end{python}
\end{minipage}

Moreover, \pilot supports the use of control structures such as \textit{if-else} and \textit{loop} statements, enabling the LLM to create dynamic feedback policies. An example of this is shown in \cref{tab:full_example}. In this sample, the LLM generates a policy function \pyth{overtake_using_right_lane}. The function first checks if there is a right lane available and a vehicle in front of the ego vehicle. If both conditions are met, it enters a loop to wait for a safe opportunity to change into the right lane. Once in the right lane, the function enters another loop to monitor the distance to the target vehicle until it has successfully overtaken the target vehicle. The \pyth{yield} statements allow the AD system to take control and execute the low-level actions required to follow the high-level policy set by the LLM.

This example demonstrates how \pilot leverages the reasoning capabilities of LLMs to generate dynamic, context-aware driving policies. By combining the strategic decision-making of LLMs with the real-time execution of classical AD algorithms, \pilot enables more flexible instruction following for AVs.
\section{\pilot Benchmark}
We introduce \pilotb, the first benchmark for evaluating the instruction-following capabilities of LLM-based agents in AD. As shown in \cref{fig:framework}, \pilotb consists of three key components: a simulator, a dataset, and an evaluator.

\subsection{Simulator}
The \pilotb simulator is built upon HighwayEnv~\cite{leurent_environment_2018}, a widely used platform for AD research and tactical decision-making. HighwayEnv offers various driving models and simulates realistic multi-vehicle interactions. We extend HighwayEnv with interfaces suitable for LLM-based agents and implement custom intersections to diversify the driving scenarios.

\subsection{Dataset}
The \pilot dataset consists of 4,900 semi-human-annotated traffic scenes, with a subset of 500 samples split as the test set. Each data sample includes:
\begin{itemize}
    \item An instruction $\Imbf$: a high-level task description.
    \item An initial state: used to initialize the simulator.
    \item Goal state criteria: aligned with the instruction $\Imbf$.
\end{itemize}

The dataset covers diverse driving scenarios, as shown in Table~\ref{tab:stat}. For each driving scenario, \pilot includes various situations. Taking turning scenarios as an example, the diversity is reflected in several variables, such as the ego vehicle's initial position and state, the specific task (turning left/right or going straight), the number of other vehicles, and their positions and states. The driving model parameters for other vehicles are randomly initialized, and each scenario is assigned a random seed.

\begin{figure}[!t]
    \centering
    \includegraphics[width=0.9\linewidth]{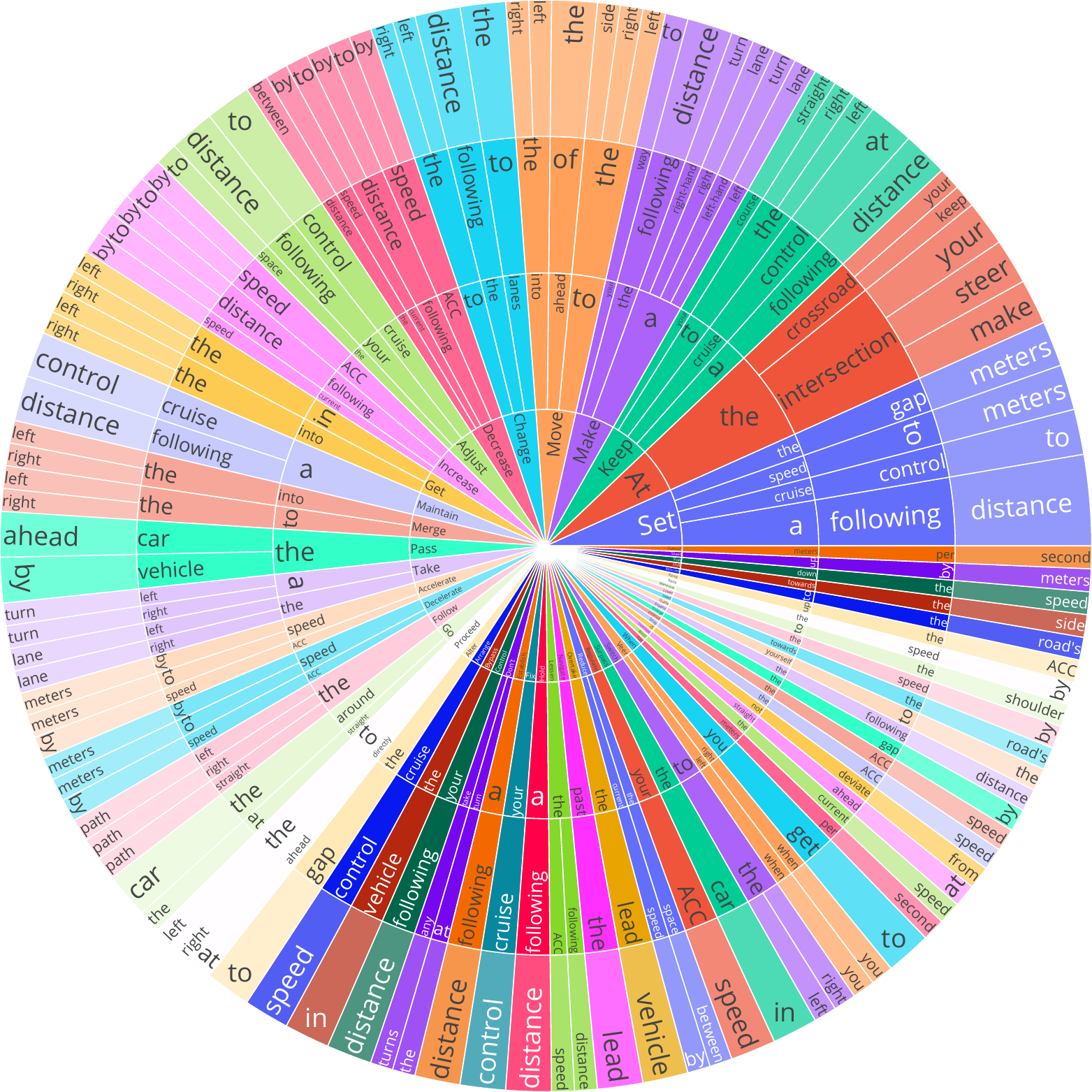}
    \caption{Distribution of instructions in the \pilotb dataset.}
    \label{fig:stat}
\end{figure}

The dataset also contains a variety of instructions reflecting realistic in-cabin human instructions categorized by maneuver types (e.g., routing, lane changing, overtaking) and scenario types (highway and intersection). \cref{fig:stat} shows the distribution of the first four words in the instructions, highlighting their diversity. Main statistics are provided in \cref{tab:stat}.

\begin{table}[!t]
    \centering
    \resizebox{0.8\linewidth}{!}{
    \begin{tabular}{l|l}
        \hline
        Statistic & Value\\
        \hline
        Total scenarios & 4,900 (100\%)\\
        \hline
        Distance-related & 1,200 (24.5\%) \\
        Speed-related & 1,200 (24.5\%) \\
        Pulling over & 200 (4.1\%) \\
        Routing & 1,500 (30.2\%) \\
        Lane changing & 400 (8.2\%)\\
        Overtaking & 400 (8.2\%)\\
        \hline
        Highway & 3,400 (69.4\%)\\
        Intersection & 1,500 (30.6\%)\\
        \hline
        Average instruction length & 7.6 words\\
        Maximum instruction length & 14 words\\
        Minimum instruction length & 2 words\\
        \hline
    \end{tabular}
    }
    \caption{Main statistics of the \pilotb dataset.}
    \vspace{-2mm}
    \label{tab:stat}
\end{table}

\subsection{Evaluator}
\label{sec:evaluator}
The \pilotb evaluator incorporates metrics to assess the safety and efficiency of the agent driving policies.

\paragraph{Safety Metric}
Time-to-collision (TTC) is used to measure the vehicle's ability to maintain safe distances and avoid collisions. In a scenario with $n+1$ vehicles (ego vehicle labeled as $0$), the TTC ($\tau_i$) with vehicle $i$ is calculated using the ego vehicle's velocity $\mathbf{v}_0$ and position $\mathbf{p}_0$, and vehicle $i$'s velocity $\mathbf{v}_i$ and position $\mathbf{p}_i$ ($1 \leq i \leq n$):

\begin{equation}
\tau_i = -\frac{(\mathbf{p}_0 - \mathbf{p}_i) \cdot (\mathbf{v}_0-\mathbf{v}_i)}{\|\mathbf{v}_0-\mathbf{v}_i\|^2}
\end{equation}

The minimum positive TTC value $\tau_{\min}$ among all $n$ vehicles over all time steps $t$ (up to task completion time $T$) represents the nearest potential collision:

\begin{equation}
\tau_{\min} = \min_{1\leq i\leq n, \, 1\leq t\leq T} \tau^{t}_{i}
\end{equation}

TTC scores are empirically assigned based on a 2-second safety margin, with values above 2 seconds considered safe and given a score of 100:

\begin{equation}
    \operatorname{TTC} =
    \begin{cases}
        100 & \text{if $\tau_{\min} > 2$} \\
        \max(0, 100 - \frac{1}{\tau_{\min}}) & \text{if $0 < \tau_{\min} \leq 2$} \\
        0 & \text{if $\tau_{\min} \leq 0$}
    \end{cases}
\end{equation}

Speed variance (SV) is considered as another safety metric~\cite{garber_factors_1989}. The ego vehicle's speed standard deviation $\sigma_{0}$ is calculated as:

\begin{equation}
\sigma_{0} = \sqrt{\frac{\sum_{t=1}^T{(\|\mathbf{v}^t_0\|-\mu_0)^2}}{T}},
\end{equation}

where $\mu_0$ is the mean speed:

\begin{equation}
\mu_{0} = \frac{\sum_{t=1}^T{\|\mathbf{v}^t_0\|}}{T}.
\end{equation}

The SV score is defined as:

\begin{equation}
\operatorname{SV} = \max(0, 100 \cdot (1 - {\sigma_{0}}/{\sigma_{\operatorname{safe}}})),
\end{equation}

where $\sigma_{\operatorname{safe}}$ is the maximum safe speed deviation, determined empirically.

\paragraph{Efficiency Metric}
The time efficiency (TE) score evaluates the policy's ability to complete the task within a predefined time limit $T_{\operatorname{limit}}$:

\begin{equation}
\operatorname{TE} = \max(0, 100 \cdot (1 - {T}/{T_{\operatorname{limit}}}))
\end{equation}

A TE score closer to 100 indicates a more efficient policy.

\paragraph{Task Completion Criteria}
A task is considered successfully completed when the agent achieves the objectives specified in the instruction while maintaining safety (i.e., avoiding collisions) and efficiency (i.e., finishing within the allotted time). For example, a lane change task is completed when the vehicle is in the target lane with its heading aligned with the lane's direction within a specified threshold.

\paragraph{Overall Scoring}
The final score aggregates all individual metrics, weighted according to their importance:
\begin{equation}
\label{eq:score}
\operatorname{Score} = W_{\operatorname{TTC}} \cdot \operatorname{TTC} + W_{\operatorname{SV}} \cdot \operatorname{SV} + W_{\operatorname{TE}} \cdot \operatorname{TE}
\end{equation}

% These evaluation metrics provide a comprehensive assessment of agents' performance in \pilotb, enabling researchers to compare approaches and identify areas for improvement in generating safe and efficient driving policies.

\begin{figure}[!t]
    \centering
    \includegraphics[width=\linewidth]{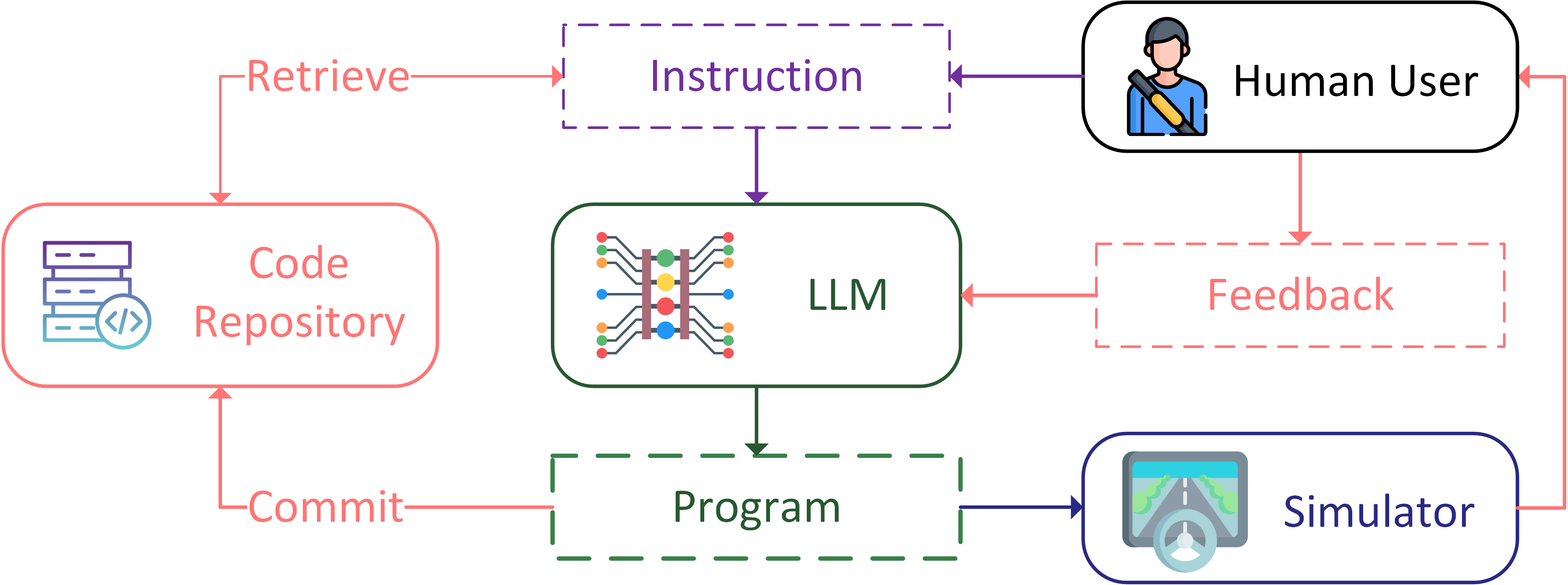}
    \caption{Schematic view of the human feedback baseline based on retrieval-augmented generation.}
    \vspace{-3mm}
    \label{fig:hf}
    % \vspace{-15pt}
\end{figure}

\begin{table*}[!t]
    \centering
    \small
    \resizebox{0.95\linewidth}{!}{
    \begin{tabular}{l|c|llllll}
        \hline
        Method & Learning & Collision ($\downarrow$) & Completion $(\uparrow)$ & TTC Score $(\uparrow)$ & SV Score $(\uparrow)$ & TE Score $(\uparrow)$ & Driving Score $(\uparrow)$\\
        \hline
        \rowcolor{Black!20} \multicolumn{8}{c}{Heuristic Baselines} \\
        IDM & \multirow{3}{3em}{N/A} & 0.0\% & 20.4\% & 92.8 & 80.9 & 71.0 & 17.3\\
        MOBIL & & 0.0\% & 15.3\% & 82.8 & 85.9 & 46.7 & 11.7\\
        Human Driver & & 0.0\% & 98.0\% & 93.4 & 86.3 & 81.3 & 84.6 \\
        \hline
        \rowcolor{Black!20} \multicolumn{8}{c}{LLM-based Agents} \\
        Llama 2 & \multirow{5}{3em}{0-Shot} 
        & 0.0\% & 20.4\% & 92.8 & 81.4 & 68.9 & 17.3\\
        PaLM 2 & 
        & 3.1\% & 35.7\% & 78.6 & 76.5 & 78.8 & 12.8 \\
        ChatGPT & 
        & 1.0\% & 40.8\% & 83.9 & 75.5 & 74.0 & 27.8\\
        GPT-4 & 
        & 4.1\% & 72.4\% & 61.3 & 70.5 & 74.0 & 28.5 \\
        GPT-4 Turbo & 
        & 3.1\% & 74.5\% & 73.2 & 70.1 & 73.9 & 39.1\\ 
        \hline
        Llama 2 & \multirow{5}{3em}{3-Shot} 
        & 1.0\% (\dec{+1.0}) & 63.3\% (\inc{+42.9}) & 68.3 (\dec{-24.5}) & 71.4 (\dec{-10.0}) & 73.8 (\inc{+4.9}) & 39.6 (\inc{+22.3}) \\
        PaLM 2 &  
        & 3.1\% (0.0) & 71.4\% (\inc{+35.7}) & 73.4 (\dec{-5.2}) & 69.7 (\dec{-6.8}) & 72.0 (\dec{-6.8}) & 36.6 (\inc{+23.8}) \\
        ChatGPT &  
        & 4.1\% (\dec{+3.1}) & 83.7\% (\inc{+42.9}) & 70.9 (\dec{-13.0}) & 73.7 (\dec{-1.8}) & 77.7 (\inc{+3.7}) & 41.2 (\inc{+13.4}) \\
        GPT-4 &  
        & 1.0\% (\inc{-3.1}) & 84.7\% (\inc{+12.3}) & 69.4 (\inc{+8.1}) & 72.0 (\inc{+1.5}) & 76.7 (\inc{+2.7}) & 55.9 (\inc{+27.4})\\
        GPT-4 Turbo &  
        & 1.0\% (\inc{-2.1}) & 80.6\% (\inc{+6.1}) & 69.1 (\dec{-4.1}) & 71.5 (\inc{+1.4}) & 74.9 (\inc{+1.0}) & 52.6 (\inc{+13.5})\\ 
        \hline
        Llama 2 & \multirow{5}{4em}{Human Feedback} 
        & 0.9\% (\inc{-0.1}) & 32.7\% (\dec{-30.6}) & 83.9 (\inc{+15.6}) & 74.6 (\inc{+3.2}) & 74.6 (\inc{+0.8}) & 21.4 (\dec{-18.2}) \\
        PaLM 2 &  
        & 0.0\% (\inc{-3.1}) & 45.5\% (\dec{-25.9}) & 81.1 (\inc{+7.7}) & 74.4 (\inc{+4.7}) & 77.6 (\inc{+5.6}) & 35.6 (\dec{-1.0})\\
        ChatGPT &  
        & 0.9\% (\inc{-3.2}) & 80.9\% (\dec{-2.8}) & 70.2 (\dec{-0.7}) & 71.7 (\dec{-2.0}) & 77.5 (\dec{-0.2}) & 54.2 (\inc{+13.0})\\
        GPT-4 &  
        & 0.9\% (\inc{-0.1}) & 92.7\% (\inc{+8.0}) & 67.6 (\dec{-1.8}) & 72.2 (\inc{+0.2}) & 76.8 (\inc{+0.1}) & 64.0 (\inc{+8.1})\\
        GPT-4 Turbo &  
        & 0.9\% (\inc{-0.1}) & 87.3\% (\inc{+6.7}) & 74.1 (\inc{+5.0}) & 73.4 (\inc{+1.9}) & 75.5 (\inc{+0.6}) & 60.5 (\inc{+7.9})\\ 
        \hline
    \end{tabular}
    }
    \caption{Comparison of baselines on \pilotb under different learning settings. Arrows ($\uparrow/\downarrow$) indicate higher or lower scores are better, respectively. Numbers in parentheses show the performance change compared to the previous setting, with (\inc{green}/\dec{red}) indicating performance \inc{improvements} and \dec{degradation}, respectively.}
    \label{tab:main-result}
\end{table*}

\section{Experiments and Results}
Our experiments are designed to establish baselines for the proposed \pilotb. The main objectives are to assess the performance of LLM-based agents in interpreting human instructions within driving contexts and to evaluate the capability of LLMs to generate code for motion planning using provided functional primitives. 

\subsection{Experiment Setup}
\paragraph{Models} We conduct benchmarking using various state-of-the-art large language models, which include both open-source and proprietary solutions\footnote{OpenAI API accessed in October and November 2023.}:
\begin{itemize}
\item \textbf{\textit{Llama 2}}~\cite{touvron_llama_2023-1} ({\ttfamily llama-2-70b-chat}) 
\item \textbf{\textit{PaLM 2}}~\cite{anil_palm_2023} ({\ttfamily code-bison}) 
\item \textbf{\textit{ChatGPT}}~\cite{openai_introducing_2023} ({\ttfamily gpt-3.5-turbo}) 
\item \textbf{\textit{GPT-4}}~\cite{openai_gpt-4_2023} ({\ttfamily gpt-4}) 
\item \textbf{\textit{GPT-4-Turbo}}~\cite{openai_gpt-4-turbo_2023} ({\ttfamily gpt-4-1106-preview}) 
\end{itemize}

\paragraph{Code Execution} To execute a LMP, the Python \pyth{exec} function is used. It takes the LMP code as an input string, along with two dictionaries that define the execution scope: (i) \pyth{apis}, which includes all APIs the code may invoke, and (ii) \pyth{policies}, an initially empty dictionary that will eventually contain a \pyth{policy} variable. If the LMP is designed to return a generator, this generator is extracted from the \pyth{policies} dictionary after the execution of the \pyth{exec} function.

\paragraph{Evaluation} We set a 60-second time limit for each scenario. If a task is not completed within this time frame, the test case is considered a failure, and the simulation is terminated, resulting in a driving score of 0. For successful cases, the driving score is calculated as follows:
\begin{equation}
\operatorname{Driving\ Score}=\frac{\alpha}{N_s}\sum_{i=1}^{N_s}\operatorname{Score}_i-\beta\cdot P_{\operatorname{collision}}
\label{eq:driving-score}
\end{equation}
Here, $N_s$ represents the number of successful test cases, $\alpha$ is the success rate (ranging from 0 to 1), $\beta$ is the collision rate (also between 0 and 1), and $P_{\text{collision}}$ is the penalty factor for collisions (set at 500 in our experiments). The $\text{Score}_i$ for each individual sample is calculated based on \cref{eq:score}, with weights $W_{\text{TTC}}$, $W_{\text{SV}}$, and $W_{\text{TE}}$ set to 0.5, 0.3, and 0.2, respectively. The driving score provides a comprehensive assessment of the agent's performance in \pilotb.

\subsection{Baselines}
\subsubsection{Heuristic Baselines}
In autonomous driving, rule-based models are favored for their deterministic and interpretable nature. In our experiments, we employ two rule-based baseline policies: the Intelligent Driver Model (IDM)\cite{treiber_congested_2000} and the Minimizing Overall Braking Induced by Lane Changes (MOBIL) principle\cite{kesting_general_2007}. IDM describes a rule for updating the acceleration of a vehicle to avoid collisions based on the proximity and relative velocity of the vehicle to the object directly in front. MOBIL is an extension of IDM where a lane change is executed if the prospective new lane offers a more favorable driving scenario and the maneuver can be conducted safely. These baselines can be considered as operating on \textit{random chance}, as the policy is independent of user instructions but follows predefined rules. Additionally, we include a \textit{human performance} baseline, where a licensed human driver controls the vehicle in the simulation using arrow keys on the keyboard, following the commands and visuals displayed. This baseline provides a reference for human-level performance on \pilotb.

\subsubsection{Zero-Shot and Few-Shot Baselines}
We also include the zero-shot and few-shot chain-of-thought (CoT) prompting~\cite{wei_chain--thought_2022} as baseline methods. The zero-shot CoT baseline leverages the prompt \textit{``How to complete the task step by step."} In the few-shot setting, we follow the standard practice~\cite{brown_language_2020} by including 3 human-written code exemplars before the test instance. These in-context examples help the model adapt to the tasks in \pilotb. The in-context examples are created by a programmer proficient in Python. They are provided with API descriptions and are allowed to write and test their code using the non-test scenarios. The same set of three examples are used for all test cases. The API Docs are also the same across all test cases for both zero-shot and few-shot settings.

\subsubsection{Human Feedback Baselines}
LLMs have demonstrated proficiency in generating coherent solutions across various tasks without requiring additional fine-tuning. However, when it comes to code generation, especially for complex scenarios, they can produce suboptimal results. The autoregressive nature of these models poses a significant challenge, as tokens generated early in a sequence cannot be modified in the same iteration. This constraint limits the models' ability to refine initial responses, potentially impacting the effectiveness of the generated code~\cite{wang_voyager_2023}.

To address these challenges and enhance the performance of LLMs on tasks in \pilotb, we introduce a human-in-the-loop approach, where the LLMs serve not only as planners but also as engines for incorporating human feedback. As shown in \cref{fig:hf}, human feedback is infused into the learning process, transforming the \pilot framework from an open-loop system into an evolving feedback loop. After an LLM generates a policy program ($\Pmbf$), it can integrate human feedback ($\Fmbf$) with context-specific guidance, enabling the LLM to refine its output.

This approach does not involve fine-tuning or training a new LLM. Instead, we leverage the RAG approach~\cite{lewis_retrieval-augmented_2020} and introduce a code repository module, which is implemented with the Chroma~\cite{chroma_ai-native_2023} vector database. This repository allows for the storage and retrieval of effective code snippets to be used in similar situations. Following Voyager~\cite{wang_voyager_2023}, we use LLM-generated function descriptions, which are then converted into vectors to serve as database keys. These keys are paired with the corresponding function codes as their values. If the human is satisfied with the policy, the code is committed to the repository for future reference. The human feedback learning process is conducted on the non-test set, involving interaction with 100 random samples for each method to ensure fair comparisons.

\subsection{Quantitative Results}
In this section, we present the experimental results on the \pilotb, summarizing the performance of various methods, including heuristic baselines, 0-shot and 3-shot baselines, and human feedback baselines. These results are shown in \cref{tab:main-result}.

\paragraph{Rule-Based Methods}
We first evaluate the performance of rule-based approaches, specifically the IDM and MOBIL algorithms. Both methods achieve a zero collision rate in \pilotb. However, it is important to note that these methods operate independently of the provided instructions. Without considering human instructions, IDM and MOBIL achieve success rates of 20.4\% and 15.3\%, respectively. These results provide a reference point for assessing the effectiveness of LLM-based agents in following human instructions. 

\paragraph{LLM-Based Agents}
We assess the effectiveness of off-the-shelf LLMs to reason in the context of autonomous driving and following human instructions. In the 0-shot setting, GPT models and PaLM 2, given only the API Docs, driving context, and instructions, achieve an obvious performance advantage compared to rule-based methods in terms of task completion. This improvement can be attributed to the LLMs' ability to understand and generate code based on natural language instructions, leveraging their extensive pretraining on diverse data sources. However, the 0-shot setting also results in an increase in the collision rate (1\%-4\%), indicating that the generated code policies do not fully capture the complexities of driving tasks.

When provided with training examples that include exemplar code (3-shot setting), all the evaluated LLMs exhibit notable improvements in completion rates. For instance, Llama 2's completion rate increases from 20.4\% in the 0-shot setting to 63.3\% in the 3-shot setting. This improvement demonstrates the effectiveness of few-shot learning in adapting LLMs to specific task domains, as the exemplar code provides a clearer understanding of the desired behavior and API usage patterns.

The integration of GPT models with human-in-the-loop learning further enhances their performance in driving tasks. Notably, GPT-4 achieves the highest driving score of $64.0$. This improvement highlights the great potential of LLMs in following instructions in the driving context when combined with human feedback. The human feedback loop allows for iterative refinement of the generated policies, enabling the LLMs to learn from their mistakes and continuously incorporate human knowledge to generate more effective driving strategies.

\subsection{Discussion}

\paragraph{Safety}
Safety is a critical factor for any AD system~\cite{wang_empowering_2024}. In our context, ensuring that the system does not execute an unsafe action, even if instructed to do so by the human, is essential for responsible autonomy. Our framework includes safety checks at two levels: (1) When the LLM generates code based on human instructions, it first reasons about the safety of the action in the context of the current driving environment. By leveraging the vast knowledge and reasoning capabilities of LLMs, the system can identify potentially unsafe instructions and generate alternative actions that prioritize safety. (2) Even if the LLM fails to catch an unsafe instruction, we have a second layer of safety checks built into the APIs that interface with the vehicle actuators. These APIs encapsulate classical planning and control algorithms with hard safety constraints. This level of safety check serves as a fail-safe mechanism, ensuring that the system does not execute actions that violate safety constraints, regardless of the LLM's output.

\paragraph{Instruction Tuning}
Instruction tuning has been validated as a technique for LLMs to acquire domain-specific knowledge efficiently~\cite{cao_maplm_2024}. Fine-tuning LLMs on driving-related tasks could enhance their performance and reduce collision rates by providing them with more targeted knowledge specific to driving. However, this may also jeopardize their generalization capability when there exist some spurious correlations in the finetuning datasets (e.g. between driving context and policy)~\cite{ye_spurious_2024}. Therefore, we hope that \pilotb can provide a platform to encourage future researchers to explore generalizable solutions to enhance LLMs' abilities in driving tasks.

\paragraph{Multimodal LLMs}
Exploring the integration of vision modality in multimodal LLMs for AD tasks is another important future direction. Our current framework relies on textual input and output, which limits its ability to perceive and understand the visual world. By incorporating visual perception and understanding capabilities, multimodal LLMs could process and reason about real-world driving scenes more effectively.

\section{Conclusion}
\label{sec:conclusion}
In this paper, we introduced \pilot, a novel framework that integrates Large Language Models (LLMs) into autonomous driving (AD) systems, enabling human-like interaction in diverse driving contexts. We also created \pilotb, a benchmark for evaluating the instruction-following capabilities of LLM-based agents in AD. Our experiments demonstrated that off-the-shelf LLMs can generate code policies for driving tasks based on human instructions. However, the notable collision rate indicates the need for further research to fully capture the complexities and safety requirements of real-world driving scenarios. \pilot provides a starting point for leveraging LLMs in solving AD-related tasks. As LLMs continue to evolve and integrate with more modalities, we envision a future where they can significantly contribute to the development of safe, efficient, and user-friendly autonomous vehicles.
\section*{Acknowledgment}
This work is partially funded by the Digital Twin Roadmap of InfoTech Labs, Toyota Motor North America. The contents of this paper only reflect the views of the authors, who are responsible for the facts and the accuracy of the data presented herein. The contents do not necessarily reflect the official views of Toyota Motor North America.
{
    \small
    \bibliographystyle{ieeenat_fullname}
    \bibliography{main,ma}

\begin{thebibliography}{66}
\providecommand{\natexlab}[1]{#1}
\providecommand{\url}[1]{\texttt{#1}}
\expandafter\ifx\csname urlstyle\endcsname\relax
  \providecommand{\doi}[1]{doi: #1}\else
  \providecommand{\doi}{doi: \begingroup \urlstyle{rm}\Url}\fi

\bibitem[Anil et~al.(2023)Anil, Dai, Firat, Johnson, Lepikhin, Passos, Shakeri, Taropa, Bailey, Chen, Chu, Clark, Shafey, Huang, Meier-Hellstern, Mishra, Moreira, Omernick, Robinson, Ruder, Tay, Xiao, Xu, Zhang, Abrego, Ahn, Austin, Barham, Botha, Bradbury, Brahma, Brooks, Catasta, Cheng, Cherry, Choquette-Choo, Chowdhery, Crepy, Dave, Dehghani, Dev, Devlin, Díaz, Du, Dyer, Feinberg, Feng, Fienber, Freitag, Garcia, Gehrmann, Gonzalez, Gur-Ari, Hand, Hashemi, Hou, Howland, Hu, Hui, Hurwitz, Isard, Ittycheriah, Jagielski, Jia, Kenealy, Krikun, Kudugunta, Lan, Lee, Lee, Li, Li, Li, Li, Li, Lim, Lin, Liu, Liu, Maggioni, Mahendru, Maynez, Misra, Moussalem, Nado, Nham, Ni, Nystrom, Parrish, Pellat, Polacek, Polozov, Pope, Qiao, Reif, Richter, Riley, Ros, Roy, Saeta, Samuel, Shelby, Slone, Smilkov, So, Sohn, Tokumine, Valter, Vasudevan, Vodrahalli, Wang, Wang, Wang, Wang, Wieting, Wu, Xu, Xu, Xue, Yin, Yu, Zhang, Zheng, Zheng, Zhou, Zhou, Petrov, and Wu]{anil_palm_2023}
Rohan Anil, Andrew~M. Dai, Orhan Firat, Melvin Johnson, Dmitry Lepikhin, Alexandre Passos, Siamak Shakeri, Emanuel Taropa, Paige Bailey, Zhifeng Chen, Eric Chu, Jonathan~H. Clark, Laurent~El Shafey, Yanping Huang, Kathy Meier-Hellstern, Gaurav Mishra, Erica Moreira, Mark Omernick, Kevin Robinson, Sebastian Ruder, Yi Tay, Kefan Xiao, Yuanzhong Xu, Yujing Zhang, Gustavo~Hernandez Abrego, Junwhan Ahn, Jacob Austin, Paul Barham, Jan Botha, James Bradbury, Siddhartha Brahma, Kevin Brooks, Michele Catasta, Yong Cheng, Colin Cherry, Christopher~A. Choquette-Choo, Aakanksha Chowdhery, Clément Crepy, Shachi Dave, Mostafa Dehghani, Sunipa Dev, Jacob Devlin, Mark Díaz, Nan Du, Ethan Dyer, Vlad Feinberg, Fangxiaoyu Feng, Vlad Fienber, Markus Freitag, Xavier Garcia, Sebastian Gehrmann, Lucas Gonzalez, Guy Gur-Ari, Steven Hand, Hadi Hashemi, Le Hou, Joshua Howland, Andrea Hu, Jeffrey Hui, Jeremy Hurwitz, Michael Isard, Abe Ittycheriah, Matthew Jagielski, Wenhao Jia, Kathleen Kenealy, Maxim Krikun, Sneha Kudugunta, Chang
  Lan, Katherine Lee, Benjamin Lee, Eric Li, Music Li, Wei Li, YaGuang Li, Jian Li, Hyeontaek Lim, Hanzhao Lin, Zhongtao Liu, Frederick Liu, Marcello Maggioni, Aroma Mahendru, Joshua Maynez, Vedant Misra, Maysam Moussalem, Zachary Nado, John Nham, Eric Ni, Andrew Nystrom, Alicia Parrish, Marie Pellat, Martin Polacek, Alex Polozov, Reiner Pope, Siyuan Qiao, Emily Reif, Bryan Richter, Parker Riley, Alex~Castro Ros, Aurko Roy, Brennan Saeta, Rajkumar Samuel, Renee Shelby, Ambrose Slone, Daniel Smilkov, David~R. So, Daniel Sohn, Simon Tokumine, Dasha Valter, Vijay Vasudevan, Kiran Vodrahalli, Xuezhi Wang, Pidong Wang, Zirui Wang, Tao Wang, John Wieting, Yuhuai Wu, Kelvin Xu, Yunhan Xu, Linting Xue, Pengcheng Yin, Jiahui Yu, Qiao Zhang, Steven Zheng, Ce Zheng, Weikang Zhou, Denny Zhou, Slav Petrov, and Yonghui Wu.
\newblock {PaLM} 2 {Technical} {Report}.
\newblock \emph{arXiv}, 2023.

\bibitem[Brown et~al.(2020)Brown, Mann, Ryder, Subbiah, Kaplan, Dhariwal, Neelakantan, Shyam, Sastry, Askell, Agarwal, Herbert-Voss, Krueger, Henighan, Child, Ramesh, Ziegler, Wu, Winter, Hesse, Chen, Sigler, Litwin, Gray, Chess, Clark, Berner, McCandlish, Radford, Sutskever, and Amodei]{brown_language_2020}
Tom~B. Brown, Benjamin Mann, Nick Ryder, Melanie Subbiah, Jared Kaplan, Prafulla Dhariwal, Arvind Neelakantan, Pranav Shyam, Girish Sastry, Amanda Askell, Sandhini Agarwal, Ariel Herbert-Voss, Gretchen Krueger, Tom Henighan, Rewon Child, Aditya Ramesh, Daniel~M. Ziegler, Jeffrey Wu, Clemens Winter, Christopher Hesse, Mark Chen, Eric Sigler, Mateusz Litwin, Scott Gray, Benjamin Chess, Jack Clark, Christopher Berner, Sam McCandlish, Alec Radford, Ilya Sutskever, and Dario Amodei.
\newblock Language {Models} are {Few}-{Shot} {Learners}.
\newblock In \emph{{NeurIPS}}, 2020.

\bibitem[Cao et~al.(2024)Cao, Zhou, Ma, Ye, Cui, Tang, Cao, Liang, Wang, Rehg, and Zheng]{cao_maplm_2024}
Xu Cao, Tong Zhou, Yunsheng Ma, Wenqian Ye, Can Cui, Kun Tang, Zhipeng Cao, Kaizhao Liang, Ziran Wang, James~M. Rehg, and Chao Zheng.
\newblock {MAPLM}: {A} {Real}-{World} {Large}-{Scale} {Vision}-{Language} {Dataset} for {Map} and {Traffic} {Scene} {Understanding}.
\newblock In \emph{{CVPR}}, 2024.

\bibitem[Chen et~al.(2024)Chen, Sinavski, Hünermann, Karnsund, Willmott, Birch, Maund, and Shotton]{chen_driving_2024}
Long Chen, Oleg Sinavski, Jan Hünermann, Alice Karnsund, Andrew~James Willmott, Danny Birch, Daniel Maund, and Jamie Shotton.
\newblock Driving with {LLMs}: {Fusing} {Object}-{Level} {Vector} {Modality} for {Explainable} {Autonomous} {Driving}.
\newblock In \emph{{ICRA}}, 2024.

\bibitem[{Chroma}(2023)]{chroma_ai-native_2023}
{Chroma}.
\newblock The {AI}-native open-source embedding database, 2023.
\newblock URL \url{https://github.com/chroma-core/chroma}.

\bibitem[Cui et~al.(2023)Cui, Yang, Zhou, Ma, Lu, Li, Chen, Panchal, and Wang]{cui_large_2023}
Can Cui, Zichong Yang, Yupeng Zhou, Yunsheng Ma, Juanwu Lu, Lingxi Li, Yaobin Chen, Jitesh Panchal, and Ziran Wang.
\newblock Large {Language} {Models} for {Autonomous} {Driving}: {Real}-{World} {Experiments}.
\newblock \emph{arXiv}, 2023.

\bibitem[Cui et~al.(2024{\natexlab{a}})Cui, Ma, Cao, Ye, and Wang]{cui_drive_2024}
Can Cui, Yunsheng Ma, Xu Cao, Wenqian Ye, and Ziran Wang.
\newblock Drive as {You} {Speak}: {Enabling} {Human}-{Like} {Interaction} with {Large} {Language} {Models} in {Autonomous} {Vehicles}.
\newblock In \emph{{WACVW}}, 2024{\natexlab{a}}.

\bibitem[Cui et~al.(2024{\natexlab{b}})Cui, Ma, Cao, Ye, and Wang]{cui_receive_2024}
Can Cui, Yunsheng Ma, Xu Cao, Wenqian Ye, and Ziran Wang.
\newblock Receive, {Reason}, and {React}: {Drive} as {You} {Say} with {Large} {Language} {Models} in {Autonomous} {Vehicles}.
\newblock \emph{IEEE Intelligent Transportation Systems Magazine}, 2024{\natexlab{b}}.

\bibitem[Cui et~al.(2024{\natexlab{c}})Cui, Ma, Cao, Ye, Zhou, Liang, Chen, Lu, Yang, Liao, Gao, Li, Tang, Cao, Zhou, Liu, Yan, Mei, Cao, Wang, and Zheng]{cui_survey_2024}
Can Cui, Yunsheng Ma, Xu Cao, Wenqian Ye, Yang Zhou, Kaizhao Liang, Jintai Chen, Juanwu Lu, Zichong Yang, Kuei-Da Liao, Tianren Gao, Erlong Li, Kun Tang, Zhipeng Cao, Tong Zhou, Ao Liu, Xinrui Yan, Shuqi Mei, Jianguo Cao, Ziran Wang, and Chao Zheng.
\newblock A {Survey} on {Multimodal} {Large} {Language} {Models} for {Autonomous} {Driving}.
\newblock In \emph{{WACVW}}, 2024{\natexlab{c}}.

\bibitem[Ding et~al.(2023)Ding, Han, Xu, Zhang, and Li]{ding_hilm-d_2023}
Xinpeng Ding, Jianhua Han, Hang Xu, Wei Zhang, and Xiaomeng Li.
\newblock {HiLM}-{D}: {Towards} {High}-{Resolution} {Understanding} in {Multimodal} {Large} {Language} {Models} for {Autonomous} {Driving}.
\newblock \emph{arXiv}, 2023.

\bibitem[Fu et~al.(2024{\natexlab{a}})Fu, Lei, Wen, Cai, Mao, Dou, Shi, and Qiao]{fu_limsim_2024}
Daocheng Fu, Wenjie Lei, Licheng Wen, Pinlong Cai, Song Mao, Min Dou, Botian Shi, and Yu Qiao.
\newblock {LimSim}++: {A} {Closed}-{Loop} {Platform} for {Deploying} {Multimodal} {LLMs} in {Autonomous} {Driving}.
\newblock \emph{arXiv}, 2024{\natexlab{a}}.

\bibitem[Fu et~al.(2024{\natexlab{b}})Fu, Li, Wen, Dou, Cai, Shi, and Qiao]{fu_drive_2024}
Daocheng Fu, Xin Li, Licheng Wen, Min Dou, Pinlong Cai, Botian Shi, and Yu Qiao.
\newblock Drive {Like} a {Human}: {Rethinking} {Autonomous} {Driving} with {Large} {Language} {Models}.
\newblock In \emph{{WACVW}}, 2024{\natexlab{b}}.

\bibitem[Gao et~al.(2024)Gao, Li, Long, Yang, and Shen]{gao_survey_2024}
Haoxiang Gao, Yaqian Li, Kaiwen Long, Ming Yang, and Yiqing Shen.
\newblock A {Survey} for {Foundation} {Models} in {Autonomous} {Driving}.
\newblock \emph{arXiv}, 2024.

\bibitem[Garber and Gadiraju(1989)]{garber_factors_1989}
Nicholas~J. Garber and Ravi Gadiraju.
\newblock Factors affecting speed variance and its influence on accidents.
\newblock \emph{Transportation research record}, 1989.

\bibitem[Grigorescu et~al.(2020)Grigorescu, Trasnea, Cocias, and Macesanu]{grigorescu_survey_2020}
Sorin Grigorescu, Bogdan Trasnea, Tiberiu Cocias, and Gigel Macesanu.
\newblock A survey of deep learning techniques for autonomous driving.
\newblock \emph{Journal of Field Robotics}, 2020.

\bibitem[Hu et~al.(2023)Hu, Yang, Chen, Li, Sima, Zhu, Chai, Du, Lin, Wang, Lu, Jia, Liu, Dai, Qiao, and Li]{hu_planning-oriented_2023}
Yihan Hu, Jiazhi Yang, Li Chen, Keyu Li, Chonghao Sima, Xizhou Zhu, Siqi Chai, Senyao Du, Tianwei Lin, Wenhai Wang, Lewei Lu, Xiaosong Jia, Qiang Liu, Jifeng Dai, Yu Qiao, and Hongyang Li.
\newblock Planning-oriented {Autonomous} {Driving}.
\newblock In \emph{{CVPR}}, 2023.

\bibitem[Huang et~al.(2023)Huang, Wang, Zhang, Li, Wu, and Fei-Fei]{huang_voxposer_2023}
Wenlong Huang, Chen Wang, Ruohan Zhang, Yunzhu Li, Jiajun Wu, and Li Fei-Fei.
\newblock {VoxPoser}: {Composable} {3D} {Value} {Maps} for {Robotic} {Manipulation} with {Language} {Models}.
\newblock In \emph{{CoRL}}, 2023.

\bibitem[Jiang et~al.(2023)Jiang, Chen, Xu, Liao, Chen, Zhou, Zhang, Liu, Huang, and Wang]{jiang_vad_2023}
Bo Jiang, Shaoyu Chen, Qing Xu, Bencheng Liao, Jiajie Chen, Helong Zhou, Qian Zhang, Wenyu Liu, Chang Huang, and Xinggang Wang.
\newblock {VAD}: {Vectorized} {Scene} {Representation} for {Efficient} {Autonomous} {Driving}.
\newblock In \emph{{ICCV}}, 2023.

\bibitem[Jin et~al.(2023)Jin, Shen, Peng, Liu, Qin, Li, Xie, Gao, Zhou, and Gong]{jin_surrealdriver_2023}
Ye Jin, Xiaoxi Shen, Huiling Peng, Xiaoan Liu, Jingli Qin, Jiayang Li, Jintao Xie, Peizhong Gao, Guyue Zhou, and Jiangtao Gong.
\newblock {SurrealDriver}: {Designing} {Generative} {Driver} {Agent} {Simulation} {Framework} in {Urban} {Contexts} based on {Large} {Language} {Model}.
\newblock \emph{arXiv}, 2023.

\bibitem[Kesting et~al.(2007)Kesting, Treiber, and Helbing]{kesting_general_2007}
Arne Kesting, Martin Treiber, and Dirk Helbing.
\newblock General {Lane}-{Changing} {Model} {MOBIL} for {Car}-{Following} {Models}.
\newblock \emph{Transportation Research Record}, 2007.

\bibitem[Leurent(2018)]{leurent_environment_2018}
Edouard Leurent.
\newblock An {Environment} for {Autonomous} {Driving} {Decision}-{Making}, 2018.
\newblock URL \url{https://github.com/Farama-Foundation/HighwayEnv}.

\bibitem[Lewis et~al.(2020)Lewis, Perez, Piktus, Petroni, Karpukhin, Goyal, Küttler, Lewis, Yih, Rocktäschel, Riedel, and Kiela]{lewis_retrieval-augmented_2020}
Patrick Lewis, Ethan Perez, Aleksandra Piktus, Fabio Petroni, Vladimir Karpukhin, Naman Goyal, Heinrich Küttler, Mike Lewis, Wen-tau Yih, Tim Rocktäschel, Sebastian Riedel, and Douwe Kiela.
\newblock Retrieval-{Augmented} {Generation} for {Knowledge}-{Intensive} {NLP} {Tasks}.
\newblock In \emph{{NeurIPS}}, 2020.

\bibitem[Li et~al.(2024)Li, Wang, Mao, Ivanovic, Veer, Leung, and Pavone]{li_driving_2024}
Boyi Li, Yue Wang, Jiageng Mao, Boris Ivanovic, Sushant Veer, Karen Leung, and Marco Pavone.
\newblock Driving {Everywhere} with {Large} {Language} {Model} {Policy} {Adaptation}.
\newblock In \emph{{CVPR}}, 2024.

\bibitem[Li et~al.(2023)Li, Bai, Cai, Wen, Fu, Zhang, Yang, Cai, Ma, Guo, Gao, Dou, Shi, Liu, He, and Qiao]{li_towards_2023}
Xin Li, Yeqi Bai, Pinlong Cai, Licheng Wen, Daocheng Fu, Bo Zhang, Xuemeng Yang, Xinyu Cai, Tao Ma, Jianfei Guo, Xing Gao, Min Dou, Botian Shi, Yong Liu, Liang He, and Yu Qiao.
\newblock Towards {Knowledge}-driven {Autonomous} {Driving}.
\newblock \emph{arXiv}, 2023.

\bibitem[Liang et~al.(2023)Liang, Huang, Xia, Xu, Hausman, Ichter, Florence, and Zeng]{liang_code_2023}
Jacky Liang, Wenlong Huang, Fei Xia, Peng Xu, Karol Hausman, Brian Ichter, Pete Florence, and Andy Zeng.
\newblock Code as {Policies}: {Language} {Model} {Programs} for {Embodied} {Control}.
\newblock In \emph{{ICRA}}, 2023.

\bibitem[Lu et~al.(2023)Lu, Peng, Cheng, Galley, Chang, Wu, Zhu, and Gao]{lu_chameleon_2023}
Pan Lu, Baolin Peng, Hao Cheng, Michel Galley, Kai-Wei Chang, Ying~Nian Wu, Song-Chun Zhu, and Jianfeng Gao.
\newblock Chameleon: {Plug}-and-{Play} {Compositional} {Reasoning} with {Large} {Language} {Models}.
\newblock In \emph{{NeurIPS}}, 2023.

\bibitem[Mao et~al.(2023{\natexlab{a}})Mao, Qian, Ye, Zhao, and Wang]{mao_gpt-driver_2023}
Jiageng Mao, Yuxi Qian, Junjie Ye, Hang Zhao, and Yue Wang.
\newblock {GPT}-{Driver}: {Learning} to {Drive} with {GPT}.
\newblock \emph{NeurIPS Workshop on Foundation Models for Decision Making}, 2023{\natexlab{a}}.

\bibitem[Mao et~al.(2023{\natexlab{b}})Mao, Ye, Qian, Pavone, and Wang]{mao_language_2023}
Jiageng Mao, Junjie Ye, Yuxi Qian, Marco Pavone, and Yue Wang.
\newblock A {Language} {Agent} for {Autonomous} {Driving}.
\newblock \emph{arXiv}, 2023{\natexlab{b}}.

\bibitem[Marcu et~al.(2023)Marcu, Chen, Hünermann, Karnsund, Hanotte, Chidananda, Nair, Badrinarayanan, Kendall, Shotton, and Sinavski]{marcu_lingoqa_2023}
Ana-Maria Marcu, Long Chen, Jan Hünermann, Alice Karnsund, Benoit Hanotte, Prajwal Chidananda, Saurabh Nair, Vijay Badrinarayanan, Alex Kendall, Jamie Shotton, and Oleg Sinavski.
\newblock {LingoQA}: {Video} {Question} {Answering} for {Autonomous} {Driving}.
\newblock \emph{arXiv}, 2023.

\bibitem[{McKinsey Center for Future Mobility}(2023)]{mckinsey_center_for_future_mobility_autonomous_2023}
{McKinsey Center for Future Mobility}.
\newblock Autonomous driving’s future: convenient and connected, 2023.
\newblock URL \url{https://www.mckinsey.com/industries/automotive-and-assembly/our-insights/autonomous-drivings-future-convenient-and-connected}.

\bibitem[Nie et~al.(2023)Nie, Peng, Wang, Cai, Han, Xu, and Zhang]{nie_reason2drive_2023}
Ming Nie, Renyuan Peng, Chunwei Wang, Xinyue Cai, Jianhua Han, Hang Xu, and Li Zhang.
\newblock {Reason2Drive}: {Towards} {Interpretable} and {Chain}-based {Reasoning} for {Autonomous} {Driving}.
\newblock \emph{arXiv}, 2023.

\bibitem[{OpenAI}(2023)]{openai_gpt-4-turbo_2023}
{OpenAI}.
\newblock {GPT}-4-{Turbo} with {128K} context and lower prices, the new {Assistants} {API}, {GPT}-4 {Turbo} with {Vision}, {DALL}·{E} 3 {API}, and more, 2023.
\newblock URL \url{https://openai.com/blog/new-models-and-developer-products-announced-at-devday}.

\bibitem[OpenAI(2023)]{openai_gpt-4_2023}
OpenAI.
\newblock {GPT}-4 {Technical} {Report}.
\newblock \emph{arXiv}, 2023.

\bibitem[{OpenAI}(2023)]{openai_introducing_2023}
{OpenAI}.
\newblock Introducing {ChatGPT}, 2023.
\newblock URL \url{https://openai.com/blog/chatgpt}.

\bibitem[Pan et~al.(2024)Pan, Yaman, Nesti, Mallik, Allievi, Velipasalar, and Ren]{pan_vlp_2024}
Chenbin Pan, Burhaneddin Yaman, Tommaso Nesti, Abhirup Mallik, Alessandro~G. Allievi, Senem Velipasalar, and Liu Ren.
\newblock {VLP}: {Vision} {Language} {Planning} for {Autonomous} {Driving}.
\newblock In \emph{{CVPR}}, 2024.

\bibitem[Sha et~al.(2023)Sha, Mu, Jiang, Chen, Xu, Luo, Li, Tomizuka, Zhan, and Ding]{sha_languagempc_2023}
Hao Sha, Yao Mu, Yuxuan Jiang, Li Chen, Chenfeng Xu, Ping Luo, Shengbo~Eben Li, Masayoshi Tomizuka, Wei Zhan, and Mingyu Ding.
\newblock {LanguageMPC}: {Large} {Language} {Models} as {Decision} {Makers} for {Autonomous} {Driving}.
\newblock \emph{arXiv}, 2023.

\bibitem[Shalev-Shwartz et~al.(2017)Shalev-Shwartz, Shammah, and Shashua]{shalev-shwartz_formal_2017}
Shai Shalev-Shwartz, Shaked Shammah, and Amnon Shashua.
\newblock On a {Formal} {Model} of {Safe} and {Scalable} {Self}-driving {Cars}.
\newblock \emph{arXiv}, 2017.

\bibitem[Shao et~al.(2023)Shao, Wang, Chen, Waslander, Li, and Liu]{shao_reasonnet_2023}
Hao Shao, Letian Wang, Ruobing Chen, Steven~L. Waslander, Hongsheng Li, and Yu Liu.
\newblock {ReasonNet}: {End}-to-{End} {Driving} with {Temporal} and {Global} {Reasoning}.
\newblock In \emph{{CVPR}}. arXiv, 2023.

\bibitem[Shao et~al.(2024)Shao, Hu, Wang, Waslander, Liu, and Li]{shao_lmdrive_2024}
Hao Shao, Yuxuan Hu, Letian Wang, Steven~L. Waslander, Yu Liu, and Hongsheng Li.
\newblock {LMDrive}: {Closed}-{Loop} {End}-to-{End} {Driving} with {Large} {Language} {Models}.
\newblock In \emph{{CVPR}}, 2024.

\bibitem[Sima et~al.(2023)Sima, Renz, Chitta, Chen, Zhang, Xie, Luo, Geiger, and Li]{sima_drivelm_2023}
Chonghao Sima, Katrin Renz, Kashyap Chitta, Li Chen, Hanxue Zhang, Chengen Xie, Ping Luo, Andreas Geiger, and Hongyang Li.
\newblock {DriveLM}: {Driving} with {Graph} {Visual} {Question} {Answering}.
\newblock \emph{arXiv}, 2023.

\bibitem[Singh et~al.(2023)Singh, Blukis, Mousavian, Goyal, Xu, Tremblay, Fox, Thomason, and Garg]{singh_progprompt_2023}
Ishika Singh, Valts Blukis, Arsalan Mousavian, Ankit Goyal, Danfei Xu, Jonathan Tremblay, Dieter Fox, Jesse Thomason, and Animesh Garg.
\newblock {ProgPrompt}: {Generating} {Situated} {Robot} {Task} {Plans} using {Large} {Language} {Models}.
\newblock In \emph{{ICRA}}, 2023.

\bibitem[Sun et~al.(2024)Sun, Zheng, Xie, Liu, Chu, Qiu, Xu, Ding, Li, Geng, Wu, Wang, Chen, Yin, Ren, Fu, He, Yuan, Liu, Liu, Li, Dong, Cheng, Zhang, Heng, Dai, Luo, Wang, Wen, Qiu, Guo, Xiong, Liu, and Li]{sun_survey_2024}
Jiankai Sun, Chuanyang Zheng, Enze Xie, Zhengying Liu, Ruihang Chu, Jianing Qiu, Jiaqi Xu, Mingyu Ding, Hongyang Li, Mengzhe Geng, Yue Wu, Wenhai Wang, Junsong Chen, Zhangyue Yin, Xiaozhe Ren, Jie Fu, Junxian He, Wu Yuan, Qi Liu, Xihui Liu, Yu Li, Hao Dong, Yu Cheng, Ming Zhang, Pheng~Ann Heng, Jifeng Dai, Ping Luo, Jingdong Wang, Ji-Rong Wen, Xipeng Qiu, Yike Guo, Hui Xiong, Qun Liu, and Zhenguo Li.
\newblock A {Survey} of {Reasoning} with {Foundation} {Models}.
\newblock \emph{arXiv}, 2024.

\bibitem[Surís et~al.(2023)Surís, Menon, and Vondrick]{suris_vipergpt_2023}
Dídac Surís, Sachit Menon, and Carl Vondrick.
\newblock {ViperGPT}: {Visual} {Inference} via {Python} {Execution} for {Reasoning}.
\newblock In \emph{{ICCV}}, 2023.

\bibitem[Tian et~al.(2024)Tian, Gu, Li, Liu, Hu, Wang, Zhan, Jia, Lang, and Zhao]{tian_drivevlm_2024}
Xiaoyu Tian, Junru Gu, Bailin Li, Yicheng Liu, Chenxu Hu, Yang Wang, Kun Zhan, Peng Jia, Xianpeng Lang, and Hang Zhao.
\newblock {DriveVLM}: {The} {Convergence} of {Autonomous} {Driving} and {Large} {Vision}-{Language} {Models}.
\newblock \emph{arXiv}, 2024.

\bibitem[Touvron et~al.(2023)Touvron, Martin, Stone, Albert, Almahairi, Babaei, Bashlykov, Batra, Bhargava, Bhosale, Bikel, Blecher, Ferrer, Chen, Cucurull, Esiobu, Fernandes, Fu, Fu, Fuller, Gao, Goswami, Goyal, Hartshorn, Hosseini, Hou, Inan, Kardas, Kerkez, Khabsa, Kloumann, Korenev, Koura, Lachaux, Lavril, Lee, Liskovich, Lu, Mao, Martinet, Mihaylov, Mishra, Molybog, Nie, Poulton, Reizenstein, Rungta, Saladi, Schelten, Silva, Smith, Subramanian, Tan, Tang, Taylor, Williams, Kuan, Xu, Yan, Zarov, Zhang, Fan, Kambadur, Narang, Rodriguez, Stojnic, Edunov, and Scialom]{touvron_llama_2023-1}
Hugo Touvron, Louis Martin, Kevin Stone, Peter Albert, Amjad Almahairi, Yasmine Babaei, Nikolay Bashlykov, Soumya Batra, Prajjwal Bhargava, Shruti Bhosale, Dan Bikel, Lukas Blecher, Cristian~Canton Ferrer, Moya Chen, Guillem Cucurull, David Esiobu, Jude Fernandes, Jeremy Fu, Wenyin Fu, Brian Fuller, Cynthia Gao, Vedanuj Goswami, Naman Goyal, Anthony Hartshorn, Saghar Hosseini, Rui Hou, Hakan Inan, Marcin Kardas, Viktor Kerkez, Madian Khabsa, Isabel Kloumann, Artem Korenev, Punit~Singh Koura, Marie-Anne Lachaux, Thibaut Lavril, Jenya Lee, Diana Liskovich, Yinghai Lu, Yuning Mao, Xavier Martinet, Todor Mihaylov, Pushkar Mishra, Igor Molybog, Yixin Nie, Andrew Poulton, Jeremy Reizenstein, Rashi Rungta, Kalyan Saladi, Alan Schelten, Ruan Silva, Eric~Michael Smith, Ranjan Subramanian, Xiaoqing~Ellen Tan, Binh Tang, Ross Taylor, Adina Williams, Jian~Xiang Kuan, Puxin Xu, Zheng Yan, Iliyan Zarov, Yuchen Zhang, Angela Fan, Melanie Kambadur, Sharan Narang, Aurelien Rodriguez, Robert Stojnic, Sergey Edunov, and Thomas
  Scialom.
\newblock Llama 2: {Open} {Foundation} and {Fine}-{Tuned} {Chat} {Models}.
\newblock \emph{arXiv}, 2023.

\bibitem[Treiber et~al.(2000)Treiber, Hennecke, and Helbing]{treiber_congested_2000}
Martin Treiber, Ansgar Hennecke, and Dirk Helbing.
\newblock Congested traffic states in empirical observations and microscopic simulations.
\newblock \emph{Physical Review E}, 2000.

\bibitem[Wang et~al.(2023{\natexlab{a}})Wang, Xie, Jiang, Mandlekar, Xiao, Zhu, Fan, and Anandkumar]{wang_voyager_2023}
Guanzhi Wang, Yuqi Xie, Yunfan Jiang, Ajay Mandlekar, Chaowei Xiao, Yuke Zhu, Linxi Fan, and Anima Anandkumar.
\newblock Voyager: {An} {Open}-{Ended} {Embodied} {Agent} with {Large} {Language} {Models}.
\newblock \emph{arXiv}, 2023{\natexlab{a}}.

\bibitem[Wang et~al.(2023{\natexlab{b}})Wang, Wang, Quan, Feng, Xu, Nie, Wang, Khabsa, Firooz, and Liu]{wang_mustie_2023}
Qifan Wang, Jingang Wang, Xiaojun Quan, Fuli Feng, Zenglin Xu, Shaoliang Nie, Sinong Wang, Madian Khabsa, Hamed Firooz, and Dongfang Liu.
\newblock {MUSTIE}: {Multimodal} {Structural} {Transformer} for {Web} {Information} {Extraction}.
\newblock In \emph{{ACL}}, 2023{\natexlab{b}}.

\bibitem[Wang et~al.(2023{\natexlab{c}})Wang, Xie, Hu, Zou, Fan, Tong, Wen, Wu, Deng, Li, Tian, Lu, Zhu, Wang, Qiao, and Dai]{wang_drivemlm_2023}
Wenhai Wang, Jiangwei Xie, ChuanYang Hu, Haoming Zou, Jianan Fan, Wenwen Tong, Yang Wen, Silei Wu, Hanming Deng, Zhiqi Li, Hao Tian, Lewei Lu, Xizhou Zhu, Xiaogang Wang, Yu Qiao, and Jifeng Dai.
\newblock {DriveMLM}: {Aligning} {Multi}-{Modal} {Large} {Language} {Models} with {Behavioral} {Planning} {States} for {Autonomous} {Driving}.
\newblock \emph{arXiv}, 2023{\natexlab{c}}.

\bibitem[Wang et~al.(2023{\natexlab{d}})Wang, Li, and Ji]{wang_code4struct_2023}
Xingyao Wang, Sha Li, and Heng Ji.
\newblock {Code4Struct}: {Code} {Generation} for {Few}-{Shot} {Event} {Structure} {Prediction}.
\newblock In \emph{{ACL}}, 2023{\natexlab{d}}.

\bibitem[Wang et~al.(2023{\natexlab{e}})Wang, Peng, Jabbarvand, and Ji]{wang_leti_2023}
Xingyao Wang, Hao Peng, Reyhaneh Jabbarvand, and Heng Ji.
\newblock {LeTI}: {Learning} to {Generate} from {Textual} {Interactions}.
\newblock \emph{arXiv}, 2023{\natexlab{e}}.

\bibitem[Wang et~al.(2024)Wang, Jiao, Lang, Zhan, Huang, Wang, Yang, and Zhu]{wang_empowering_2024}
Yixuan Wang, Ruochen Jiao, Chengtian Lang, Sinong~Simon Zhan, Chao Huang, Zhaoran Wang, Zhuoran Yang, and Qi Zhu.
\newblock Empowering {Autonomous} {Driving} with {Large} {Language} {Models}: {A} {Safety} {Perspective}.
\newblock In \emph{{ICLR} {Workshop} on {LLM} {Agents}}, 2024.

\bibitem[{Wayve Technologies Ltd}(2023)]{wayve_technologies_ltd_lingo-1_2023}
{Wayve Technologies Ltd}.
\newblock {LINGO}-1: {Exploring} {Natural} {Language} for {Autonomous} {Driving}, 2023.
\newblock URL \url{https://wayve.ai/thinking/lingo-natural-language-autonomous-driving/}.

\bibitem[Wei et~al.(2022)Wei, Wang, Schuurmans, Bosma, Ichter, Xia, Chi, Le, and Zhou]{wei_chain--thought_2022}
Jason Wei, Xuezhi Wang, Dale Schuurmans, Maarten Bosma, Brian Ichter, Fei Xia, Ed Chi, Quoc Le, and Denny Zhou.
\newblock Chain-of-{Thought} {Prompting} {Elicits} {Reasoning} in {Large} {Language} {Models}.
\newblock In \emph{{NeurIPS}}, 2022.

\bibitem[Wen et~al.(2023)Wen, Yang, Fu, Wang, Cai, Li, Ma, Li, Xu, Shang, Zhu, Sun, Bai, Cai, Dou, Hu, Shi, and Qiao]{wen_road_2023}
Licheng Wen, Xuemeng Yang, Daocheng Fu, Xiaofeng Wang, Pinlong Cai, Xin Li, Tao Ma, Yingxuan Li, Linran Xu, Dengke Shang, Zheng Zhu, Shaoyan Sun, Yeqi Bai, Xinyu Cai, Min Dou, Shuanglu Hu, Botian Shi, and Yu Qiao.
\newblock On the {Road} with {GPT}-{4V}(ision): {Early} {Explorations} of {Visual}-{Language} {Model} on {Autonomous} {Driving}.
\newblock \emph{arXiv}, 2023.

\bibitem[Wen et~al.(2024)Wen, Fu, Li, Cai, Ma, Cai, Dou, Shi, He, and Qiao]{wen_dilu_2024}
Licheng Wen, Daocheng Fu, Xin Li, Xinyu Cai, Tao Ma, Pinlong Cai, Min Dou, Botian Shi, Liang He, and Yu Qiao.
\newblock {DiLu}: {A} {Knowledge}-{Driven} {Approach} to {Autonomous} {Driving} with {Large} {Language} {Models}.
\newblock In \emph{{ICLR}}, 2024.

\bibitem[Xu et~al.(2023)Xu, Zhang, Xie, Zhao, Guo, Wong, Li, and Zhao]{xu_drivegpt4_2023}
Zhenhua Xu, Yujia Zhang, Enze Xie, Zhen Zhao, Yong Guo, Kwan-Yee~K. Wong, Zhenguo Li, and Hengshuang Zhao.
\newblock {DriveGPT4}: {Interpretable} {End}-to-end {Autonomous} {Driving} via {Large} {Language} {Model}.
\newblock \emph{arXiv}, 2023.

\bibitem[Yang et~al.(2024{\natexlab{a}})Yang, Liu, Wu, Yang, Fung, Li, Huang, Cao, Wang, Wang, Ji, and Zhai]{yang_if_2024}
Ke Yang, Jiateng Liu, John Wu, Chaoqi Yang, Yi~R. Fung, Sha Li, Zixuan Huang, Xu Cao, Xingyao Wang, Yiquan Wang, Heng Ji, and Chengxiang Zhai.
\newblock If {LLM} {Is} the {Wizard}, {Then} {Code} {Is} the {Wand}: {A} {Survey} on {How} {Code} {Empowers} {Large} {Language} {Models} to {Serve} as {Intelligent} {Agents}.
\newblock \emph{ICLR Workshop on LLM Agents}, 2024{\natexlab{a}}.

\bibitem[Yang et~al.(2024{\natexlab{b}})Yang, Zhang, Li, Marta, Batool, and Folkesson]{yang_human-centric_2024}
Yi Yang, Qingwen Zhang, Ci Li, Daniel~Simões Marta, Nazre Batool, and John Folkesson.
\newblock Human-{Centric} {Autonomous} {Systems} {With} {LLMs} for {User} {Command} {Reasoning}.
\newblock In \emph{{WACVW}}, 2024{\natexlab{b}}.

\bibitem[Yang et~al.(2023)Yang, Jia, Li, and Yan]{yang_llm4drive_2023}
Zhenjie Yang, Xiaosong Jia, Hongyang Li, and Junchi Yan.
\newblock {LLM4Drive}: {A} {Survey} of {Large} {Language} {Models} for {Autonomous} {Driving}.
\newblock \emph{arXiv}, 2023.

\bibitem[Yao et~al.(2023)Yao, Zhao, Yu, Du, Shafran, Narasimhan, and Cao]{yao_react_2023}
Shunyu Yao, Jeffrey Zhao, Dian Yu, Nan Du, Izhak Shafran, Karthik Narasimhan, and Yuan Cao.
\newblock {ReAct}: {Synergizing} {Reasoning} and {Acting} in {Language} {Models}.
\newblock In \emph{{ICLR}}, 2023.

\bibitem[Ye et~al.(2024)Ye, Zheng, Cao, Ma, Hu, and Zhang]{ye_spurious_2024}
Wenqian Ye, Guangtao Zheng, Xu Cao, Yunsheng Ma, Xia Hu, and Aidong Zhang.
\newblock Spurious {Correlations} in {Machine} {Learning}: {A} {Survey}.
\newblock \emph{arXiv}, 2024.

\bibitem[Yuan et~al.(2024)Yuan, Sun, Omeiza, Zhao, Newman, Kunze, and Gadd]{yuan_rag-driver_2024}
Jianhao Yuan, Shuyang Sun, Daniel Omeiza, Bo Zhao, Paul Newman, Lars Kunze, and Matthew Gadd.
\newblock {RAG}-{Driver}: {Generalisable} {Driving} {Explanations} with {Retrieval}-{Augmented} {In}-{Context} {Learning} in {Multi}-{Modal} {Large} {Language} {Model}.
\newblock \emph{arXiv}, 2024.

\bibitem[Zan et~al.(2023)Zan, Chen, Zhang, Lu, Wu, Guan, Wang, and Lou]{zan_large_2023}
Daoguang Zan, Bei Chen, Fengji Zhang, Dianjie Lu, Bingchao Wu, Bei Guan, Yongji Wang, and Jian-Guang Lou.
\newblock Large {Language} {Models} {Meet} {NL2Code}: {A} {Survey}.
\newblock In \emph{{ACL}}, 2023.

\bibitem[Zhou et~al.(2023)Zhou, Liu, Zagar, Yurtsever, and Knoll]{zhou_vision_2023}
Xingcheng Zhou, Mingyu Liu, Bare~Luka Zagar, Ekim Yurtsever, and Alois~C. Knoll.
\newblock Vision {Language} {Models} in {Autonomous} {Driving} and {Intelligent} {Transportation} {Systems}.
\newblock \emph{arXiv}, 2023.

\bibitem[Zhou et~al.(2024)Zhou, Huang, Bu, Zeng, Li, Qiu, Zhu, Guo, Qiao, and Li]{zhou_embodied_2024}
Yunsong Zhou, Linyan Huang, Qingwen Bu, Jia Zeng, Tianyu Li, Hang Qiu, Hongzi Zhu, Minyi Guo, Yu Qiao, and Hongyang Li.
\newblock Embodied {Understanding} of {Driving} {Scenarios}.
\newblock \emph{arXiv}, 2024.

\end{thebibliography}
}

% \input{sec/suppl}

\end{document}